\documentclass[runningheads,a4paper]{llncs}

\usepackage{amssymb}
\usepackage{graphicx}
\usepackage{amsmath}

\usepackage{textpos}

\usepackage{url}

\renewcommand{\u}{\mathbf{u}}
\newcommand{\x}{\mathbf{x}}
\newcommand{\xl}[1]{\x^{(#1)}}

\newcommand{\y}{\mathbf{y}}
\newcommand{\W}{\mathbf{W}}
\newcommand{\Wh}{\hat{\W}}
\newcommand{\Win}{\W_{in}}
\newcommand{\Wout}{\W_{out}}

\newcommand{\R}{\mathbb{R}}
\newcommand{\Wl}{\Wh^{(l)}}

\newcommand{\sil}{\omega_{il}}
\renewcommand{\sin}{\omega_{in}}

\begin{document}

\mainmatter  % start of an individual contribution

\title{Reservoir Topology in Deep Echo State Networks}

\titlerunning{Reservoir Topology in Deep Echo State Networks}

\author{Claudio Gallicchio \and Alessio Micheli}

\institute{Department of Computer Science, University of Pisa,\\
Largo B. Pontecorvo, 3, 56127 Pisa, Italy\\
\path{gallicch@di.unipi.it},\path{micheli@di.unipi.it}
}

\maketitle

% --- This paper has been accepted and published
\begin{textblock*}{150mm}(-2cm,-7cm)
\noindent
\emph{This is a preprint version of the following paper, published in the proceedings of ICANN 2019:}\\
Gallicchio C., Micheli A. (2019) Reservoir Topology in Deep Echo State Networks. In: Tetko I., Kůrková V., Karpov P., Theis F. (eds) Artificial Neural Networks and Machine Learning – ICANN 2019: Workshop and Special Sessions. ICANN 2019. Lecture Notes in Computer Science, vol 11731. Springer, Cham
\end{textblock*}
%--------------------------------------------------

\begin{abstract}
Deep Echo State Networks (DeepESNs) recently extended the applicability of Reservoir Computing (RC) methods towards the field of deep learning. In this paper we study the impact of constrained reservoir topologies in the architectural design of deep reservoirs, through numerical experiments on several RC benchmarks. The major outcome of our investigation is to show the remarkable effect, in terms of predictive performance gain, achieved by the synergy between a deep reservoir construction and a structured organization of the recurrent units in each layer. Our results also indicate that a particularly advantageous architectural setting is obtained in correspondence of DeepESNs where reservoir units are structured according to a permutation recurrent matrix.

\end{abstract}

\section{Introduction}

Reservoir Computing (RC) \cite{Lukosevicius2009,Verstraeten2007experimental} delineates a class of Recurrent Neural Network (RNN) models based on the idea of separating the non-linear dynamical component of the network, i.e. the recurrent hidden reservoir layer, from the feed-forward linear readout layer.
The reservoir is initialized randomly under stability constraints and then is left untrained, leaving the burden of training to fall only on the readout part of the architecture, hence resulting in a strikingly efficient approach to RNN design.
In this context, the Echo State Network (ESN) model \cite{Jaeger2004,Jaeger2001} is a popular realization of the RC paradigm based on implementing the reservoir in terms of a discrete-time non-linear dynamical system. 
Being featured by untrained dynamics, ESNs represent an important tool to understand and characterize the operation and potentialities of recurrent neural models. Shaping the reservoir architecture in order to achieve desired properties and  optimized performance in applications, even in the absence of training of the recurrent connections, 
is %then 
one of the key goals of RC research \cite{Gallicchio2018randomized}.

In this paper we bring together two major trends in the area of ESN architectural studies.
The first one focuses on the pattern of connectivity among the recurrent units. In this case, the aim is to constrain the  random reservoir initialization process towards topologies that determine specific algebraic properties of the resulting recurrent weight matrices. A relevant class of reservoir variants in this regard is given by ESNs with orthogonal recurrent matrices \cite{White2004,Hajnal2006}, which were shown to lead to improved performance with respect to random reservoirs both in terms of memorization skills and in terms of predictive performance on non-linear tasks. In particular, reservoirs whose structure is based on permutation matrices represent particularly appealing instances of orthogonal ESNs \cite{Hajnal2006,Strauss2012}, entailing a simple and very sparse pattern of connectivity among the recurrent units. 
Other relevant architectural variants are given by reservoirs structured according to a ring topology or to form a chain of units \cite{Rodan2011,Strauss2012}. 
The second major  line of research that we consider regards the construction of hierarchically structured reservoir models. While initial studies in this context focused on composing multiple ESN modules 
to form ad-hoc architectures 
\cite{jaeger2007discovering,triefenbach2010phoneme}, recent works started analyzing the effects of stacking multiple untrained reservoir layers with the introduction of the DeepESN model \cite{Gallicchio2017DeepESN}.
On the one hand, the analysis of DeepESN dynamics contributes to uncover the intrinsic computational properties of deep neural networks in the temporal domain \cite{Gallicchio2017DeepESN,Gallicchio2018local}. On the other hand, a proper architectural design of deep reservoirs might have a huge impact in real-world applications \cite{Gallicchio2018design}, enabling effective multiple time-scales processing and at the same time preserving the training efficiency typical of RC models.

In this paper we analyze
the impact on the predictive performance given by a constrained
reservoir topology in DeepESNs. Specifically,
we consider deep architectures in which the individual reservoir layers are implemented based on permutation matrices, as well as on ring and on chain topologies. Our study is conducted in comparison to shallow ESN counterparts through numerical experiments on several benchmarks in the RC area.

The rest of this paper is structured as follows.
The DeepESN model is introduced in Section~\ref{sec.DeepESN}, while the investigated reservoir topologies 
are described in Section~\ref{sec.topology}. The experimental analysis 
is reported in Section~\ref{sec.experiments}. Finally, Section~\ref{sec.conclusions} draws conclusions and delineates future research directions.

\section{Deep Echo State Networks}
\label{sec.DeepESN}
\begin{figure}[bth]
  \centering
	  \includegraphics[width=1\textwidth]{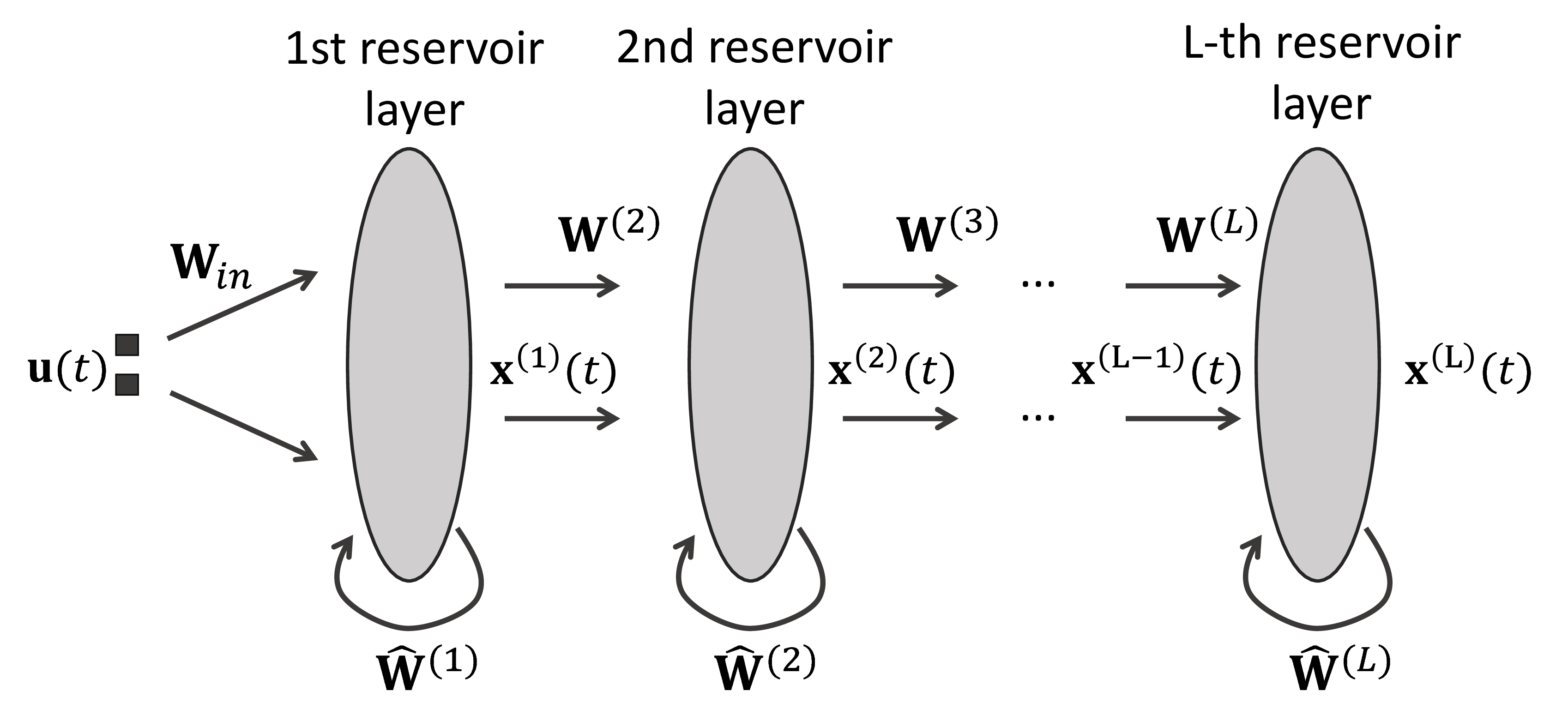}
	  \caption{Hierarchical reservoir architecture in a DeepESN.}
  \label{fig.deepreservoir}
\end{figure}

A DeepESN is an RC model in which the reservoir part is organized into a stacked composition of multiple untrained recurrent hidden layers. The external input is propagated only to the first reservoir layer, while each successive level in the deep architecture is fed by the output of the previous one, as graphically illustrated in Figure~\ref{fig.deepreservoir}. 

To fix our notation,  we use $L$ to indicate the number of layers in the deep reservoir, while $N_U$ and $N_Y$ respectively denote the sizes of input and output spaces. 
For the sake of simplicity in the presentation of the DeepESN model, here we make the assumption that all the reservoir layers are featured by the same number of units, indicated by $N_R$.
The operation of each reservoir layer can be described in terms of a discrete-time non-linear dynamical system,
whose state update equation is given in the form of an iterated mapping. In particular, at time-step $t$, the state of the first layer, i.e. $\xl{1}(t) \in \R^{N_R}$, is computed as follows:
\begin{equation}
\label{eq.layer1}
\xl{1}(t) = \tanh(\Win \u(t) + \Wh^{(1)} \xl{1}(t-1)),
\end{equation}
while the state of each successive layer $l>1$, i.e. $\xl{l}(t) \in \R^{N_R}$, is given by:
\begin{equation}
\label{eq.layerl}
\xl{l}(t) = \tanh(\W^{(l)} \xl{l-1}(t) + \Wh^{(l)} \xl{l}(t-1)).
\end{equation}
Here, $\tanh$ indicates the element-wise application of the hyperbolic tangent non-linearity, $\u(t) \in \R^{N_U}$ represents the external input at time-step $t$, while $\Win$, $\W^{(l)}$ and $\Wh^{(l)}$ respectively denote the input weight matrix (that modulates the external input stimulation to the first layer), the inter-layer weight matrix for layer $l$ (that modulates the strength of the connections from layer $l-1$ to layer $l$), and the recurrent reservoir weight matrix for layer $l$. In both the above equations~\ref{eq.layer1} and \ref{eq.layerl} we omitted the bias terms for the ease of notation. The interested reader can find in \cite{Gallicchio2017DeepESN} a more detailed description of the deep reservoir equations, framed in the more general context of leaky integrator reservoir units. In order to set up initial conditions for the state update 
equations~\ref{eq.layer1} and \ref{eq.layerl}, at time-step $0$ all reservoir layers are set to a null state, i.e. $\x^{(l)}(0) = \mathbf{0}$ for all $l = 1,\ldots, L$. 
Given this framework, it is worth noticing that a standard shallow ESN model can be seen as a special case of DeepESN in which a single reservoir layer is considered, i.e. $L = 1$.

As in standard RC approaches, the parameters of the entire reservoir component, i.e. the elements in all the weight matrices in equations~\ref{eq.layer1} and \ref{eq.layerl}, are left untrained after initialization subject to stability constraints. These are required to avoid the system dynamics to fall into unstable regimes, which would make them unsuitable for robust processing of time-series data. In the context of ESNs, the analysis of asymptotic stability is usually described in terms of the Echo State Property (ESP) \cite{Jaeger2001,Lukosevicius2009}, providing simple algebraic conditions for the initialization of reservoir weight matrices that have been recently extended to cope with the case of deep reservoirs in \cite{Gallicchio2017echo}.
Under a practical view-point, the analysis in \cite{Gallicchio2017echo} suggests to carefully control the spectral radius
of all the reservoir weight matrices in the deep reservoir. In this paper, we use $\rho^{(l)}$ to denote the spectral radius in layer $l$, i.e. the largest among the absolute values of the eigenvalues of $\Wh^{(l)}$.
A simple initialization procedure for the reservoir of a DeepESN then consists in
choosing the elements in $\Wh^{(l)}$ randomly from a uniform distribution on $[-1,1]$, subsequently re-scaling them to 
achieve desired values of $\rho^{(l)}$, typically not above unity. Similarly, the elements in $\Win$ and those in $\W^{(l)}$ (for $l>1$) are initialized randomly from a uniform distribution on $[-1,1]$, and then are re-scaled to control 
the input scaling hyper-parameter $\omega_{in} = \|\Win\|_2$, and the set of inter-layer scaling hyper-parameters 
$\omega_{il}^{(l)} = \|\W^{(l)}\|_2$.

The output of the DeepESN is computed by a simple readout tool, which linearly combines the reservoir representations developed in all the layers of the deep architecture. In formulas, the output at time-step $t$, denoted as $\y(t) \in \R^{N_Y}$, is computed by the following equation:
\begin{equation}
\label{eq.readout}
\y(t) = \Wout  \, [\x^{(1)}(t); \ldots ; \x^{(L)}(t)],
\end{equation}
where $\Wout$ is the output weight matrix, and $[\x^{(1)}(t); \ldots ; \x^{(L)}(t)]$ represents the global deep reservoir state at time-step $t$, expressed as the concatenation of all the states in the architecture.
The elements in $\Wout$ represent the only learnable weights of the DeepESN, and are typically adjusted to fit a training set by exploiting non-iterative training algorithms as in the case of standard RC models \cite{Lukosevicius2009}.
Notice that, although different patterns of reservoir-to-readout connectivity are possible \cite{Pascanu2014}, the one employed here, where all reservoir layers are used to feed the readout, has the advantage to allow the training algorithms to modulate and exploit differently the variety of representations provided by the different levels in the deep architecture.

A more comprehensive description of the characteristics and advantages of the DeepESN approach can be found in \cite{Gallicchio2018Layering}, while a constantly updated overview on the advancements achieved in this research field is given in \cite{Gallicchio2017survey}.
To date, software implementations of the DeepESN model are made publicly available as libraries for 
Python\footnote{\url{https://github.com/lucapedrelli/DeepESN}}, 
Matlab\footnote{\url{https://it.mathworks.com/matlabcentral/fileexchange/69402-deepesn}} and 
Octave\footnote{\url{https://github.com/gallicch/DeepESN_octave}}.

\section{Reservoir Topology}
\label{sec.topology}
\begin{figure}[bth]
  \centering
	  \includegraphics[width=0.9\textwidth]{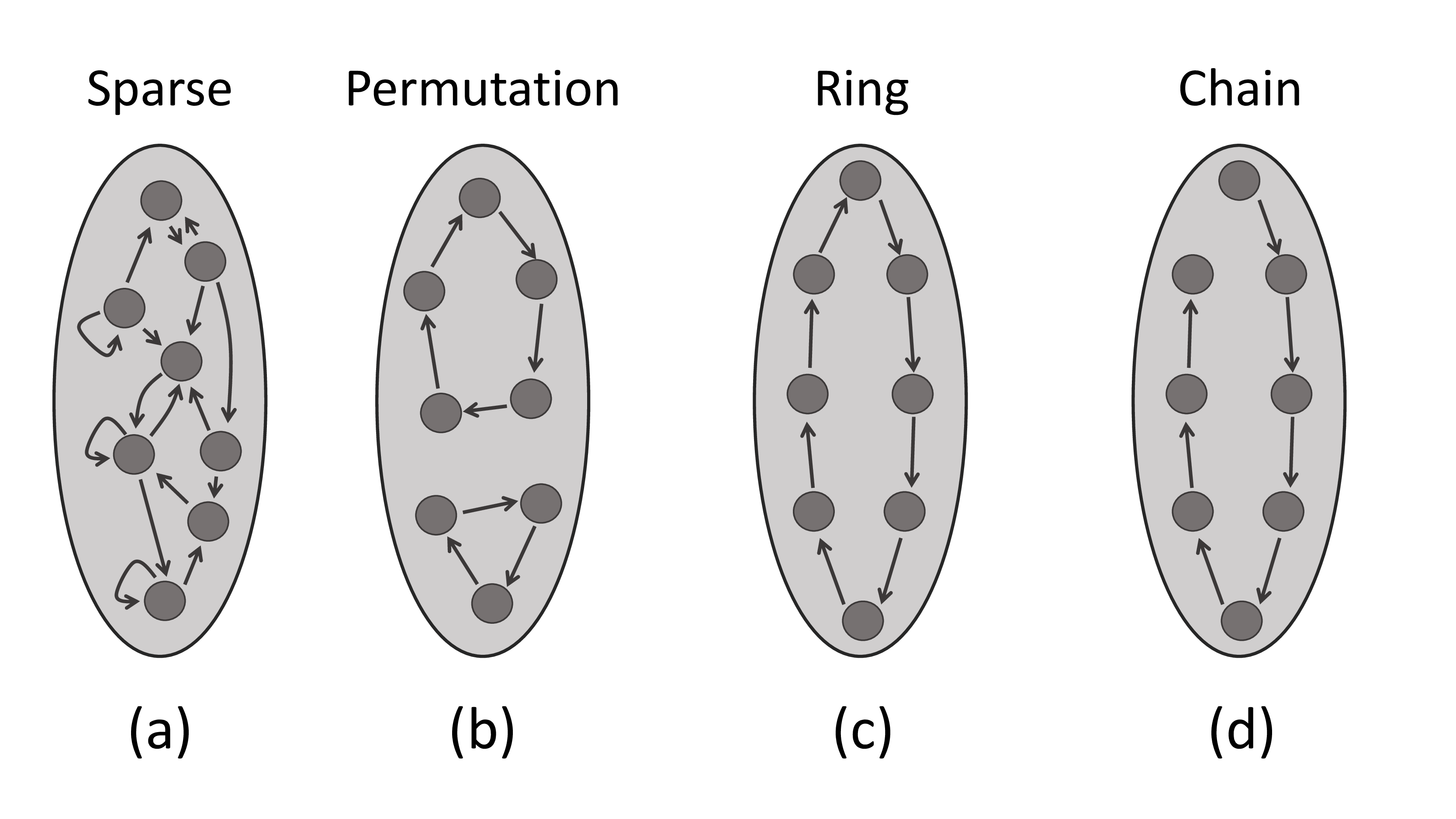}
	  \caption{Reservoir topologies of DeepESN layers.}
  \label{fig.topology}
\end{figure}

We consider DeepESN architectural variants where the recurrent weight matrix in each layer $l$, i.e. $\Wh^{(l)}$, is characterized by a specific structure, according to the topologies described in the following.
The resulting patterns of reservoir connectivity are graphically exemplified in Figure~\ref{fig.topology}.

\begin{description}
\item[Sparse] Each reservoir unit is randomly connected to a subset of the others, determining a \emph{sparse} recurrent matrix $\Wh^{(l)}$ (see Figure~\ref{fig.topology}(a)). This corresponds to a common setting used in RC practice and serves here as baseline for our analysis.
\item[Permutation] The structure of the recurrence matrix $\Wh^{(l)}$ is given by a \emph{permutation} matrix $\mathbf{P}$, i.e. we have:
\begin{equation}
\label{eq.permutation}
\Wh^{(l)} = \lambda \, \mathbf{P},
\end{equation} 
where $\mathbf{P}$ is obtained by randomly permuting the columns of the identity matrix, and $\lambda$ is a multiplicative constant that specifies the value of the non-zero recurrent weights. In this case, the spectral radius of $\Wl$ is determined by the value of $\lambda$, i.e. $\rho^{(l)} = \lambda$.
The permutation  topology implies that each row and each column of the recurrence matrix have exactly one non-zero element, resulting into a reservoir architecture that presents a variable number of disjoint cyclic structures, as graphically exemplified in Figure~\ref{fig.topology}(b). 
The levels in the deep reservoir architecture are allowed to employ different permutations, i.e. the number of cycles in each reservoir layer can be different.

In the context of shallow ESNs, this kind of topology has been empirically studied 
in \cite{Boedecker2009}, where it was shown to achieve good memorization skills at the same time improving the performance of randomly initialized reservoirs in tasks involving non-linear mappings. Interestingly, the permutation topology has been investigated in \cite{Hajnal2006} as a way to implement orthogonal reservoir matrix structures, under the name of Critical ESNs.

\item[Ring] The reservoir units  are organized to form a single \emph{ring}, as shown in Figure~\ref{fig.topology}(c). Accordingly, the recurrent weight matrix $\Wl$ is expressed as:
\begin{equation}
\label{eq.ring}
\Wl = \lambda \,
\left[
\begin{array}{c c c c}
0 & 0 & \ldots & 1\\
1 & 0 & \ldots & 0\\
\vdots & \ddots & \ddots & \vdots\\
0 & \ldots & 1 & 0\\
\end{array}
\right],
\end{equation}
where $\lambda$ is the value of non-zero recurrent weights, and determines the spectral radius of $\Wl$, i.e. $\rho^{(l)} = \lambda$. The ring topology can be easily seen as a special case of the permutation topology, where the pattern of reservoir connectivity is ruled by the specific permutation matrix in equation~\ref{eq.ring}, and the reservoir units are all part of the same cyclic structure. 

Reservoirs following this architectural organization have been subject of several studies in literature on shallow RC. Notable instances in this regard are given by the work in \cite{Strauss2012}, in which the ring topology is studied in the context of orthogonal reservoir structures, and by the work in \cite{Rodan2011}, 
where the study is carried out under the perspective of architectural design simplification for minimum complexity ESN construction. One interesting outcome of previous analysis on the ring topology is that, compared to randomly initialized reservoirs, it shows superior memory capacity that, at least in the linear case, approaches the optimal value \cite{Rodan2011}. 
While this optimal memory characterization has been extensively analyzed in literature for the more general class of orthogonal recurrent weight matrices (see e.g. \cite{White2004,Jaeger2001short,Farkas2016}), the ring topology presents the advantage of a strikingly simple (and sparse) dynamical network construction.

\item[Chain] The recurrent units are arranged in a pipeline, where each unit - except for the first one - receives in input the activation of the previous one, forming a \emph{chain} as in the example in Figure~\ref{fig.topology}(d). 
The only non-zero elements in $\Wl$ are located in the lower sub-diagonal, i.e. we have:
\begin{equation}
\label{eq.chain}
\Wl = \lambda \,
\left[
\begin{array}{c c c c}
0 & 0 & \ldots & 0\\
1 & 0 & \ldots & 0\\
\vdots & \ddots & \ddots & \vdots\\
0 & \ldots & 1 & 0\\
\end{array}
\right],
\end{equation}
where as in previous cases $\lambda$ identifies the value of non-zero weights. 
Although in this case $\Wl$ is nilpotent and hence its spectral radius is always $0$, we still operate on %the value of 
$\lambda$ to control the magnitude of recurrent weights. As such, with a slight abuse of notation, also in this case we set $\rho^{(l)} = \lambda$. 
Overall, the chain topology results in a particularly simple design strategy that, from the architectural perspective, applies a further simplification to the ring topology by removing one of the connections between the internal units. 

Literature works on shallow ESN models pointed out the merits of reservoir organizations based on a chain topology (also called delay-line reservoirs), as a very simple approach to the architectural design of the network, resulting in a model that is easier to analyze \cite{White2004} and that leads to comparable or even better performance than standard ESNs \cite{Rodan2011,Strauss2012}.
\end{description}

\section{Experiments}
\label{sec.experiments}
In this section we illustrate the experimental analysis conducted in this paper.
Specifically, in Section~\ref{sec.settings} we detail the datasets considered and the experimental settings 
adopted in our work, whereas in Section~\ref{sec.results} we report and discuss the achieved numerical results.

\subsection{Datasets and Experimental Settings}
\label{sec.settings}

In our experiments, we considered benchmark datasets featured by univariate time-series (i.e., $N_U = N_Y = 1$).

The first dataset is obtained from a non-linear auto-regressive moving average system of the 10-th order (NARMA10).
At each time-step, the input $u(t)$ comes from a uniform distribution over $[0, 0.5]$, whereas the corresponding target output
$y_{tg}(t)$ is given by the following relation:
\begin{equation}
\label{eq.narma10}
y_{tg}(t) = 0.3 \, y_{tg}(t-1)+0.05 \, y_{tg}(t-1) \sum\limits_{i = 1}^{10} y_{tg}(t-i) 
                 + 1.5 \, u(t-10) \, u(t-1)+0.1.
\end{equation}

The second dataset that we considered is the Santa Fe Laser time-series \cite{Weigend2018time}, where the input values $u(t)$
are sampled intensities from a far-infrared laser in chaotic regime, re-scaled by a factor of 0.01.
We used the Laser dataset to define a next-step prediction task, where $y_{tg}(t) = u(t+1)$ for each time-step $t$.

The last two datasets are instances of the Mackey-Glass (MG) \cite{Mackey1977oscillation,Farmer1982chaotic} time-series, obtained by discretizing the following non-linear differential equation:
\begin{equation}
\label{eq.mg}
\frac{\delta u(t)}{\delta t} = \frac{0.2 \, u(t-\tau)}{1+u(t-\tau)^{10}} - 0.1 \, u(t),
\end{equation}
where $\tau$ is a parameter of the system influencing its dynamical behavior.
We generated two MG time-series using $\tau = 17$ (MG17) and $\tau = 30$ (MG30), representing cases with increasingly complex chaotic behavior. In both cases, the elements of the time-series where shifted by -1 and then passed through the $\tanh$ squashing function as in \cite{Jaeger2004,Jaeger2001}. The two MG time-series allowed us to set up two next-step prediction tasks, where $y_{tg}(t) = u(t+1)$ for each time-step $t$. 

For NARMA10, MG17 and MG30 we generated datasets with 10000 time-steps, while the Laser dataset contained a number of 10092 samples. In all the cases, the available data was split into a training set, comprising the first 5000 time-steps, and a test set, comprising the remaining samples. The first 100 time-steps were used as transient to wash out the initial conditions. The performance of the considered RC models was evaluated in terms of mean squared error (MSE) in all the tasks.

In our experiments, we considered DeepESNs with a total number of $500$ recurrent reservoir units,
distributed evenly across the layers of the deep architecture, varying the number of layers $L$ from 2 to 5\footnote{With the only exception of the case $L=3$, where the first two layers contained 167 units and the last one contained 166 units.}.
All the reservoir layers in the deep architecture shared the same values for the scaling hyper-parameters $\rho$ and $\sil$, i.e. $\rho = \rho^{(1)} = \ldots = \rho^{(L)}$ and $\sil = \sil^{(2)} = \ldots = \sil^{(L)}$.
To account for sparsity, each reservoir unit was randomly connected to 5 units in the previous layer and to 5 units in the same layer. Of course, when considering permutation, ring and chain reservoir topologies, the connectivity of the reservoir units in each layer followed the corresponding specific structure described in Section~\ref{sec.topology}. In all the cases, we used fully-connected input weight matrices.
For every task and choice of the reservoir topology, the DeepESN hyper-parametrization was chosen by model selection on a validation set comprising the last 1000 time-steps of the training split. To this end, we performed a random search with 50 networks configurations for each number of layers, sampling the value of $\rho$ from a uniform distribution in $(0.1, 1]$, and the values of $\sin$ and $\sil$ from uniform distributions in $(0.1, 2]$. The achieved results were averaged on 10 network guesses for each hyper-parametrization explored, and readout training was performed by using pseudo-inversion.
Finally, our experimental analysis was conducted in comparison with shallow ESN setups, considering the same reservoir topologies investigated in the DeepESN case, and using the same experimental setting described above, with the only crucial exception that all the available reservoir units were organized into a single layer (i.e. $L = 1$).

\subsection{Results}
\label{sec.results}
The test MSE values obtained by DeepESNs in correspondence of all the considered types of layer-wise reservoir topology are reported in Table~\ref{tab.results}. For the sake of comparison, the same table shows the results achieved by shallow ESNs under the same architectural conditions examined in the deep case. In all the cases, the sparse reservoir topology is considered as a baseline setup for our analysis.

\begin{table}[tb]
  \centering
  \begin{tabular}{l l  l l}
    \multicolumn{4}{c}{\textbf{NARMA10}}  \\
    \hline
    \textbf{Topology} $\quad$ & \textbf{ESN}  & \textbf{DeepESN} & Layers\\
  	Sparse & $1.658 \; 10^{-4} \, (3.367 \; 10^{-5}) \quad$  & $1.647 \; 10^{-4} \,(3.415 \; 10^{-5})$ & 2\\
    Permutation & $1.354 \; 10^{-4} \, (1.589 \; 10^{-5})$  & $\underline{1.243 \; 10^{-4}} \,(1.464 \; 10^{-5})$ & 2\\
    Ring & $1.494 \; 10^{-4} \, (1.547 \; 10^{-5})$  & $1.482 \; 10^{-4} \,(1.713 \; 10^{-5})$ & 2\\
    Chain & $1.571 \; 10^{-4} \, (1.780 \; 10^{-5})$  & $1.569 \; 10^{-4} \,(2.594 \; 10^{-5})$ & 2\\
    \\
  \end{tabular}
  \vspace{3mm}
  \begin{tabular}{l l  l l}
    \multicolumn{4}{c}{\textbf{Laser}}  \\
    \hline
    \textbf{Topology} $\quad$ & \textbf{ESN}  & \textbf{DeepESN} & Layers\\
  	Sparse & $1.226 \; 10^{-3} \, (1.037 \; 10^{-4}) \quad$  & $8.228 \; 10^{-4} \,(2.309 \; 10^{-4})$ & 5\\
    Permutation & $1.312 \; 10^{-3} \, (1.385 \; 10^{-4})$  & $\underline{6.633 \; 10^{-4}} \,(8.861 \; 10^{-5})$ & 5\\
    Ring & $1.161 \; 10^{-3} \, (7.541 \; 10^{-5})$  & $7.640 \; 10^{-4} \,(4.331 \; 10^{-5})$ & 4\\
    Chain & $9.496 \; 10^{-4} \, (1.183 \; 10^{-4})$  & $8.555 \; 10^{-4} \,(8.302 \; 10^{-5})$ & 4\\

  \end{tabular}
  \vspace{3mm}
  \begin{tabular}{l l  l l}
    \multicolumn{4}{c}{\textbf{MG17}}  \\
    \hline
    \textbf{Topology} $\quad$ & \textbf{ESN}  & \textbf{DeepESN} & Layers\\
  	Sparse & $3.739 \; 10^{-9} \, (1.387 \; 10^{-9}) \quad$  & $2.328 \; 10^{-9} \,(8.299 \; 10^{-10})$ & 2\\
    Permutation & $3.093 \; 10^{-9} \, (3.241 \; 10^{-10})$  & $\underline{4.576\; 10^{-10}} \,(6.280 \; 10^{-10})$ & 5\\
    Ring & $1.585 \; 10^{-9} \, (2.989 \; 10^{-10})$  & $5.043 \; 10^{-10} \,(3.891 \; 10^{-10})$ & 5\\
    Chain & $1.950 \; 10^{-9} \, (3.745 \; 10^{-10})$  & $4.913 \; 10^{-10} \,(2.535 \; 10^{-10})$ & 3\\
    \\
  \end{tabular}
  \vspace{3mm}
  \begin{tabular}{l l  l l}
    \multicolumn{4}{c}{\textbf{MG30}}  \\
    \hline
    \textbf{Topology} $\quad$ & \textbf{ESN}  & \textbf{DeepESN} & Layers\\
  	Sparse & $1.476 \; 10^{-8} \, (1.781 \; 10^{-9}) \quad$  & $1.172 \; 10^{-8} \,(1.406 \; 10^{-9})$ & 2\\
    Permutation & $1.027 \; 10^{-8} \, (5.412 \; 10^{-10})$  & $\underline{8.618\; 10^{-9}} \,(1.457 \; 10^{-9})$ & 3\\
    Ring & $1.197 \; 10^{-8} \, (1.549 \; 10^{-9})$  & $1.078 \; 10^{-8} \,(2.066 \; 10^{-9})$ & 5\\
    Chain & $1.086 \; 10^{-8} \, (9.519 \; 10^{-10})$  & $9.096 \; 10^{-9} \,(1.803 \; 10^{-9})$ & 3\\

    \\
  \end{tabular}
  \caption{Test MSE (and std) achieved by shallow ESN and DeepESN settings 
  for different choices of the reservoir topology. 
  The last column reports the number of layers selected for DeepESN.  
  Best results for each task are underlined.
  }
\label{tab.results}  
\end{table}

The performance values reported in Table~\ref{tab.results} allow us to draw several lines of observations.
First of all, our results confirm the goodness of the considered reservoir architectural variants already in the shallow setup,
showing improved performance (i.e., a smaller MSE) with respect to the sparse baseline in all the cases (with the sole exception of permutation shallow reservoirs on Laser). 
Second, 
we observe that the performance of DeepESN with constrained topology (i.e. permutation, ring and chain) enhances that one of sparse DeepESN in all the considered tasks (with the only exception of deep reservoirs with chain architecture on Laser). Moreover, we can see that DeepESN improves the results of shallow ESN in all the tasks and for all the choices of reservoir topology, both in the constrained architectural cases and for the base sparse reservoir setup.
Taken together, results in Table~\ref{tab.results} clearly indicate the performance advantage arising from the synergy between deep organization and constrained topology as factors of architectural design of reservoirs. 
Giving a structure to the architecture of reservoirs both at a coarser level, i.e. organizing the recurrent units into layers, and at a finer level, i.e. organizing individual layers' units into cyclic or chain structures, 
amplifies the benefits brought by the two factors individually.
Finally, we notice that the best performing architecture in our experiments is the DeepESN with permutation reservoir topology, which obtained the smallest errors on all the tasks, and is put forward here as a particularly effective (yet sparse and efficient) approach to the architectural design of reservoir models.

\section{Conclusions}
\label{sec.conclusions}
In this paper we have investigated the role of reservoir topology in the architectural design of DeepESNs. 
Specifically, we focused on analyzing the effects of constraining the recurrent weight matrix of each layer according to permutation, ring and chain topologies. 
Numerical results on several RC benchmarks pointed out a striking beneficial effect arising from the combination of a deep reservoir construction with a structured organization of the recurrent units in each layer. Our results indicate that DeepESN with reservoir units arranged to obey a permutation scheme (i.e., forming multiple rings) provides a particularly advantageous design strategy for reservoirs, leading to the best performance in all the explored tasks.

While already giving interesting empirical evidences on the potentialities of deep RC architectures, the study presented in this paper opens the way to several directions for further research.
First of all, the experimental analysis described here suggests that the use of simplified deep RC models has a great potential that can be exploited massively in real-world applications. Leveraging the parsimonious design approach resulting from structured sparsity of reservoir units, the class of deep neural models studied in this work seems an ideal candidate e.g. for embedding advanced learning capabilities on resource-constrained computing devices.
On the methodological side, a natural extension of the work in this paper is to analyze the effect of a broader pool of reservoir architectural variants, including e.g. small-world \cite{Kawai2019small}, cycles with regular jumps \cite{Rodan2012} and concentric \cite{bacciu2018concentric} reservoirs.
Moreover, future research could pursue even further the simplification of architectural construction in deep RC models, reducing the impact of randomness in the network initialization in the same vein as the works on minimum complexity ESNs \cite{Rodan2011,Rodan2012}. Simplifying the reservoir structure locally to each layer can also be exploited from a more theoretically-oriented perspective, easing the mathematical analysis of dynamical properties naturally emerging in deep RNNs. In this concern, it is certainly interesting to extend fundamental mathematical results,
e.g. pertaining to short-term memory capacity \cite{Rodan2011,White2004,Jaeger2001short}, or 
to  approximation properties \cite{grigoryeva2018echo} of shallow reservoirs to the case of DeepESN.
In addition to this, we believe that the role of orthogonality in deep reservoirs, studied in this paper in relation to the individual layers of the architecture, is an intriguing concept that deserves to be investigated also at the level of global (instead of local) DeepESN dynamics. Finally, the advantages of constrained DeepESN architectures delineated in this paper can be extended to larger classes of models, including e.g. deep RC for complex data structures \cite{gallicchio2019deep}, as well as fully trained deep RNNs.

\bibliographystyle{splncs04}
\bibliography{references}

\end{document}